# Modeling of Self-sustained Neuron Population without External Stimulus

İhsan Ertuğrul Karakaş[1], Özden Özel[1], İlkay Ulusoy[1], Orhan Murat Koçak[2]

[1] Middle East Technical University

[2] Başkent University

## Abstract

Self-sustained neural activity in the absence of ongoing external input is a fundamental feature of nervous system dynamics, yet the conditions under which it can emerge in biophysically grounded network models remain incompletely understood. We studied whether a recurrent network of Hodgkin-Huxley neurons with spike-timing-dependent plasticity and intrinsic stochasticity can maintain autonomous activity after brief transient stimulation. The simulated network comprised 200 neurons (160 excitatory, 40 inhibitory) with 80% connection probability, incorporating excitatory and inhibitory STDP, probabilistic vesicle release, probabilistic synapse formation, receptor variability, and voltage-dependent inhibition. After a brief 200 ms initialization stimulus to 30 excitatory neurons, the network received no further external input. In one 1800 s simulation and two additional 500 s simulations, the network maintained sparse, irregular activity without ongoing drive. In the 1800 s run, 67% of neurons exhibited mean firing rates below 1 Hz, the population mean firing rate was 1.13 ± 1.34 Hz, participation increased across longer observation windows, and population-mean Fano factors remained near 1–2, consistent with irregular spike timing. Raster activity also showed spontaneous qualitative reorganizations in collective firing patterns over time. These findings suggest that recurrent Hodgkin-Huxley networks with plastic and stochastic synapses can sustain long-duration autonomous activity in a sparse firing regime after brief initialization.

## I. Introduction

The nervous system is remarkable not only for its responsiveness to external stimuli, but also for its ability to sustain ongoing internal activity, preserve traces of past inputs, and function under pervasive biological noise. Understanding how these properties coexist remains a central challenge in computational neuroscience. Mathematical modeling provides a primary framework for addressing this problem, yet the choice of model class is consequential. Phenomenological models such as leaky integrate-and-fire and Izhikevich neurons are computationally efficient, but their discontinuous reset rules depart from the continuous biophysical evolution of real membrane states. By contrast, biophysical models, particularly the Hodgkin-Huxley formalism, preserve the ionic mechanisms of excitability and offer a more suitable foundation for studying persistent and mechanistically interpretable neural dynamics.

This distinction is especially important when the problem is framed in terms of memory and traceability. Neural systems preserve traces of prior perturbations while simultaneously generating ongoing endogenous activity. Because this intrinsic activity evolves continuously and is shaped by recurrent interactions, refractoriness, and stochastic fluctuations, identifying the effect of a past input becomes an inverse problem in a noisy dynamical system. Although extensive work has addressed neural dynamics, plasticity, stochasticity, and input-output relations, long-duration self-sustained activity remains comparatively underexplored. Much of the literature assumes continuous external drive, focuses on brief persistence, or relies on reduced models that simplify away the conductance-based and refractory mechanisms likely to be essential for genuine autonomous activity.

Previous studies indicate that self-sustained firing depends on several interacting conditions. Recurrent loop structure is critical because refractory dynamics require activity to circulate across subpopulations rather than remain uniformly active. Network persistence also depends on a precise balance between excitation and inhibition: too little excitation leads to quiescence, whereas too much produces saturation or runaway activation. Conductance-based models appear more favorable than current-based formulations for maintaining stable persistent activity, yet extended investigations remain limited. In addition, most prior work has not treated intrinsic noise as a constitutive feature of the system, despite the probabilistic nature of vesicle release, synapse formation, receptor abundance, and spike-mediated transmission.

The present study addresses this gap by asking whether a recurrent neuronal population can sustain autonomous activity over an extended time window after only a brief initiating perturbation. To do so, we employ a Hodgkin-Huxley-based network model in which spikes emerge from continuous ionic membrane dynamics, and we combine this framework with spike-timing-dependent plasticity and biologically grounded stochastic processes. These stochastic components include probabilistic vesicle release, probabilistic synapse formation, receptor variability, and non-deterministic transmission efficacy. The network is activated by a transient current injection within the first second and then evolves without further external input for up to 30 minutes.

By integrating realistic membrane biophysics, synaptic plasticity, recurrent feedback, and intrinsic stochasticity within a single framework, this work provides a more faithful in silico platform for examining extended self-sustained neural activity. More broadly, it advances the study of how autonomous firing, memory traces, and inverse inference emerge in living neural systems whose dynamics are simultaneously recurrent, adaptive, and noisy.

## II. Literature Summary

Mathematical modeling represents one of the foundational approaches for studying the physiology of the nervous system. Given the inherent structural and functional complexity of the brain, biological processes are often expressed through mathematically tractable abstractions. Across modeling traditions, membrane dynamics constitute a unifying theme, although their formulation varies according to both the spatial-temporal scale of interest and the degree of biological abstraction. Consistent with the classification proposed by Dayan and Abbott (2001) and later elaborated by Gerstner et al. (2014), neural models may be grouped into three broad categories: biophysical, phenomenological, and population-level models. Phenomenological models, such as the leaky integrate-and-fire and Izhikevich formulations, are widely used because of their computational efficiency; however, their reliance on discontinuous reset mechanisms renders them less compatible with continuous biophysical flows, particularly in the context of well-posed inverse problems. In contrast, biophysical models, most notably the Hodgkin–Huxley formalism, aim to preserve a higher degree of biological realism by explicitly representing the ionic mechanisms underlying membrane excitability. At the single-neuron scale, such models are typically formulated within a dynamical systems framework. At the mesoscopic level, neural mass models, including the Wilson–Cowan model, describe the mean activity of large neuronal populations and treat the brain as a continuous medium, often conceptualized as a neural field. Within this hierarchical dynamical perspective, membrane potential evolution provides the primary layer of description, upon which synaptic processes are subsequently defined. This gives rise to a layered representation in which first-order membrane dynamics underpin higher-order synaptic dynamics and plasticity. Such a hierarchy is consistent with the neuron modeling framework presented by Izhikevich (2007) and with synaptic modeling approaches such as those developed by Morrison et al. (2008).

Neural cells maintain their biological individuality while forming highly organized collective structures that support learning. Through communication across multicellular networks, they encode and transmit time-dependent information about the state of the system. What fundamentally distinguishes the nervous system from other biological tissues, however, is its ability to preserve the traces, or engrams, produced by external inputs such that they remain available for subsequent retrieval.

A key issue is whether the effects of external perturbations can be reliably traced within the system, an inherently nontrivial task. The nervous system possesses ongoing intrinsic dynamics that persist independently of incoming stimuli. These dynamics are also substantially noisy due to the stochastic character of neural processes. Because the system state evolves continuously and is subject to stochastic fluctuations, isolating the trace of a given input becomes difficult. This difficulty motivates the formulation of the problem as an inverse problem in dynamical systems, where the objective is to infer the input, or cause, from the observed output.

A substantial body of literature has examined neural system dynamics across multiple spatial and temporal scales, as well as across different levels of mechanistic abstraction. In parallel, extensive research has addressed inverse problems, synaptic plasticity, stochasticity, and input–output relationships in neural systems. The interrelations among these domains have also been investigated in numerous studies. Nevertheless, one aspect that remains comparatively underexplored is the fact that the nervous system is a living system, capable of maintaining ongoing activity even in the absence of externally imposed stimuli. Possibly because the nervous system is often interpreted primarily through the lens of adaptation, its dynamical behavior is typically studied under conditions in which external input is assumed to be present. Consequently, studies explicitly addressing self-sustained neural activity remain relatively limited.

Several investigations have suggested that self-sustained activity in a neural network requires the presence of at least one loop structure. Such recurrence is considered necessary in part because of the refractory period, which prevents all neurons from remaining active simultaneously and instead imposes alternating phases of activity and silence across different neuronal subpopulations. In this context, Liu et al. (2021) reported that loop-containing architectures require intermediate neurons associated with the feedback vertex set (FVS), and that larger self-sustaining networks generally require a greater number of such neurons. At the same time, reducing the number of FVS elements effectively shortens or disrupts long feedback loops, thereby altering the network's capacity to maintain persistent activity. Conversely, if a loop is excessively short, it may also fail to support self-sustained dynamics, since neuronal activations can become temporally aligned within the same period, preventing the sequential propagation needed for persistent circulation of activity.

Other studies have shown that biologically realistic neural networks can maintain moderate levels of ongoing activity without either quiescence or saturation, provided that an appropriate balance between excitatory and inhibitory influences is preserved. When this balance is perturbed in favor of excitation, the network may transition toward runaway activity or saturation, as reported by Beggs and Plenz (2003). Although the literature often notes that receptor and neurotransmitter abundances may favor glutamatergic signaling, the spatial organization of inhibitory elements is frequently biased toward GABAergic control. The possible contribution of this structural asymmetry to self-sustained activity, however, has not yet been fully examined in a direct manner.

Beggs and Plenz derived their conclusions from multielectrode recordings obtained from mature organotypic cultures and acute cortical slices, based on approximately 70 hours of recording. However, thalamic inputs were incorporated into their framework to evoke spontaneous-like activity. Because externally injected currents were applied within a one-second interval, it is difficult to interpret the recorded activity as purely spontaneous intrinsic dynamics. In a related but distinct line of work, Teramae and Fukai (2007) investigated noise-induced synchronization in pulse-coupled oscillator networks. Importantly, their analysis focused on super-threshold neurons capable of firing in the absence of external input, thereby modeling self-sustained firing activity directly at the single-neuron level within the network.

In another study, Vogels and Abbott (2005) demonstrated sustained network activity over a one-second interval and showed that such activity is difficult to maintain under current-based formulations of excitatory–inhibitory balance, whereas conductance-based formulations provide a more suitable framework for simulating self-sustained dynamics. Their analysis was based on the leaky integrate-and-fire (LIF) model. They further reported that self-sustained activity was associated with sparse spiking, with mean firing rates of approximately 10 Hz.

None of the aforementioned studies explicitly incorporates intrinsic noise into the analysis of self-sustained activity. Yet intrinsic noise is a fundamental feature of nervous system dynamics, arising primarily from the probabilistic nature of many underlying biological mechanisms. For example, firing responses in orientation columns are commonly described in statistical terms and may be characterized by approximately Gaussian response distributions under suitable stimulus conditions. Likewise, spontaneous vesicle release at the presynaptic terminal is inherently probabilistic, as are synapse formation and the distribution and number of receptors at synaptic sites. Even action potential transmission is probabilistic in effect: although spiking substantially increases the likelihood of information transfer, it does not ensure it with absolute certainty. Consequently, a system grounded in such probabilistic mechanisms must be regarded as intrinsically stochastic.

Demonstrating self-sustained activity in a neuronal population within a digital environment is an important prerequisite for in silico experiments intended to approximate in vivo conditions. Well-developed computational models can provide insight into complex phenomena that are difficult to investigate using purely biological methods. However, direct evaluations of self-sustained neural activity over extended periods—such as days or weeks—in the complete absence of external input remain largely unavailable. The present study, therefore, aimed to examine how the physiological properties of a neuronal population evolve over a relatively long time window (500-1800 s) when no external input is provided beyond an initial current injection to 30 random excitatory neurons within the first 1s used to trigger network activity.

To achieve a high degree of biological realism, spike generation was modeled using the Hodgkin–Huxley (HH) formalism. Because the HH model exhibits a comparatively longer refractory period, simulated neurons require more time to recover between spikes, making it more difficult to establish the continuous recurrent feedback necessary for self-sustained oscillatory activity. In addition, unlike simplified models that primarily track membrane voltage or impose artificial threshold-reset rules, the HH framework explicitly represents the time-dependent dynamics of voltage-gated ion channels, including sodium and potassium conductances. Incorporating these channel dynamics provides a more mechanistic basis for synaptic adaptation and enables a biologically grounded implementation of spike-timing-dependent plasticity (STDP). To our knowledge, no previous study has directly examined plasticity in a neuronal population that maintains self-sustained activity over an extended period without external input.

A further motivation for the present work is that intrinsic noise has not been explicitly incorporated in most previous studies of self-sustained neural activity. This omission is significant because the nervous system is inherently noisy due to the probabilistic nature of many of its underlying mechanisms. In the present model, several such biologically inspired stochastic processes were incorporated. These include spontaneous vesicle release from presynaptic terminals, probabilistic synapse formation, and probabilistic variation in the number of receptors within synapses. Even action potential transmission is not strictly deterministic; rather, spiking substantially increases the probability of information transfer without guaranteeing it. Accordingly, stochasticity is treated here as an intrinsic feature of the modeled neural system rather than as an external perturbation.

## III. Method

Building on the preceding discussion of self-sustainability, stochasticity, and biological realism in neural modeling, we investigated whether a recurrent neuronal network with biologically plausible composition could sustain activity autonomously after transient stimulation. Specifically, we simulated a network of 200 neurons, of which 160 were excitatory and 40 were inhibitory, reflecting the approximately 80% excitatory and 20% inhibitory proportion reported in biological systems. The aim was to assess whether self-sustained firing could emerge from internal network dynamics alone, once the network had been initiated by a short external current pulse.

### 1) Neuron Model

The neuronal model employed in this study is based on the Hodgkin-Huxley (HH) formalism. Each neuron was represented by two functional compartments: a soma and an axonal propagation component. The soma represents the main neuronal body in which membrane potential dynamics are generated, whereas the axonal component propagates somatic voltage signals to the synaptic terminals with a neuron-specific transmission delay, thereby approximating variability in axonal length.

Membrane potential arises from ionic concentration differences across the cell membrane, which behaves electrically as a capacitor. Ionic fluxes through membrane channels alter this potential over time. Accordingly, the temporal evolution of membrane voltage was modeled as a function of membrane capacitance and the total transmembrane current, as given by

$$C_m \, du/dt = -(I_{Na} + I_K + I_L) + I_{syn} + I_{ext} \quad (1)$$

Where $I_{Na}$ and $I_K$ denote the sodium and potassium currents mediated by voltage-gated ion channels, $I_L$ denotes the leak current, $I_{syn}$ represents the net synaptic current arising from presynaptic inputs, and $I_{ext}$ denotes the externally injected current used to initiate network activity at the beginning of the simulation. Here, $C_m$ is the membrane capacitance and $du/dt$ is the rate of change of membrane potential with respect to time.

The principal complexity of the HH model lies in the voltage-dependent dynamics of its ion channels. For each ionic current, the current magnitude depends on both the effective conductance of the relevant channel population and the driving force, defined as the difference between the membrane potential and the corresponding reversal potential. For sodium and potassium channels, conductance is further modulated by gating variables that represent the fraction of channels in permissive states.

The sodium current was modeled as

$$I\_Na = g\_Na\ (u - E\_Na)\ m^3\ h$$

where *g_Na* is the maximal sodium conductance, *E_Na* is the sodium reversal potential, $m^3$ represents the probability that the three activation gates are simultaneously open, and h represents the probability that the inactivation gate is open. Similarly, the potassium current was defined as

$$I\_K = g\_K\ (u - E\_K)\ n^4$$

where *g_K* is the maximal potassium conductance, *E_K* is the potassium reversal potential, and $n^4$ denotes the probability that the four potassium activation gates are simultaneously open. The leak current was modeled in the standard linear form using the leak conductance g_L and the leak reversal potential E_L.

The core regulatory mechanism of the HH model is governed by the gating variables *m, h*, and *n*, which determine the fraction of ion channels in conductive states. These variables evolve according to voltage-dependent transition rates between open and closed states. For a generic gating variable *x* in {*m, h, n*}, the temporal dynamics were expressed as

$$dx/dt = (1/tau\_x)\ [alpha\_x\ (1 - x) - beta\_x\ x]$$

where *alpha_x* denotes the activation rate, *beta_x* denotes the deactivation rate, (*1 - x*) represents the fraction of gates in the closed state, and *x* represents the fraction in the open state. The coefficient *tau_x* was introduced to allow independent adjustment of the time scale of each gating process. The voltage-dependent functions *alpha_x* and *beta_x* were computed separately for each gating variable. The full mathematical specification of these functions, together with the complete set of HH parameters, is provided in the Appendix.

Finally, the somatic membrane potential of each neuron was transmitted to its outgoing synapses with a neuron-specific delay, thereby modeling heterogeneous axonal propagation times and approximating variability in axonal length across the network.

### 2) Synapse Model

Synapses between neurons facilitate signal transmission through the release of excitatory and inhibitory neurotransmitter vesicles from presynaptic to postsynaptic neurons. When certain events happen, a presynaptic neuron can release neurotransmitters, which bind to receptors and affect the conductance of ionic channels in the postsynaptic neuron, causing depolarization or hyperpolarization. When there is enough depolarization in the postsynaptic neuron, this can lead to an action potential, which can then transmit vesicles to other neurons downstream, causing a chain reaction, which is the activity we observe in the system.

There are two events that lead to vesicle release by a presynaptic neuron into a synapse: the arrival of an action potential or the synapse reaching a spontaneous release window with a varying period. In the first case, if the axon end of the presynaptic neuron detects a new action potential, it either releases *N_AP* vesicles with probability *p_AP_release*, or it doesn't release any vesicles. This probabilistic release mechanism, which has a probability of 0.5 for each action potential, mirrors trends observed in biological neurons in observational studies. If there is no action potential in the presynaptic neuron, but the presynaptic neuron enters a spontaneous release window, either *N_not_AP* vesicles are released into the synapse with probability *1-p_spontaneous_AP_release*, or *N_AP* vesicles are released into the synapse with probability *p_spontaneous_AP_release*, which was to 0.001 in our tests. The period between spontaneous releases gets larger with increasing presynaptic activation potential rates, simulating a reduction in spontaneous releases due to a reduction in available vesicles to release. *N_AP* and *N_not_AP* are variables drawn from a Gaussian distribution with means *mean_N_AP*, *mean_N_not_AP,* and variances *var_N_AP* and *var_N_not_AP*.

Every vesicle that has been released gets added to a pool of accumulated vesicles in that synapse, which decays exponentially with time, simulating reuptake of vesicles and preventing saturation. Each accumulated vesicle in the synapse causes a current to flow through the membrane of the postsynaptic neuron by increasing the conductance of ion channels connected to receptors on the receiving neuron, causing depolarization or polarization. The synaptic current resulting from this mechanism on the postsynaptic neuron is calculated at each step with formulas dependent on the type of the presynaptic neuron. For synapses with excitatory presynaptic neurons, the excitatory current is calculated as:

$$I\_AMPA(t) = R\_AMPA * N(t) * du\_per\_ves * g\_AMPA$$

For synapses with inhibitory presynaptic neurons, inhibitory current is calculated as:

$$I\_GABA(t) = -1 * R\_GABA * N(t) * du\_per\_ves * g\_GABA$$

where R_GABA and R_AMPA are GABA and AMPA receptors of the postsynaptic neuron for that specific synapse, N(t) is the number of accumulated vesicles at that timestep, du_per_ves is a constant used to adjust voltage change to those observed in vivo, and g_GABA and g_AMPA are the GABA and AMPA conductances, respectively.

The decay of vesicles happens exponentially at each step, according to the following formula:

$$N(t+dt) = (N(t)+new\ vesicles)*e^{(-decay\_rate*dt)}$$

where we used 100/s as the decay rate. This high decay rate ensures stable operation of the synapse even under high rates of firing.

In biological neurons, inhibitory effectiveness depends on membrane voltage: inhibition is strongest at depolarized potentials and weakens when the neuron is hyperpolarized, primarily because shunting inhibition requires excitatory currents to short-circuit. Rather than implementing full conductance-based synapses, we approximated voltage-dependent inhibition through an attenuation mechanism. To approximate this voltage dependence without full conductance-based modeling, we applied an attenuation factor to inhibitory currents when postsynaptic membrane potential fell below -70 mV. The attenuation strength increased exponentially as the neuron became more hyperpolarized:

If *u_post < -70 mV* and *I_total < 0*, then:

$$attenuation = exp(-500 \times (-70 - u\_post))$$

$$I\_GABA = I\_GABA \times attenuation$$

This exponential attenuation can reduce inhibitory efficacy up to 12.5-fold at very hyperpolarized potentials. This approximation captures the essential voltage-dependent behavior of biological shunting inhibition while maintaining computational efficiency appropriate for our network's sparse firing regime.

### 3) Spike Timing Dependent Plasticity

Our model implements biologically inspired plasticity, called spike timing dependent plasticity, that allows synaptic weights, in our case number of receptors, to be modified based on the relative timing between presynaptic and postsynaptic neurons of a synapse. We used a learning window of 50ms, meaning learning only occurs for presynaptic and postsynaptic action potentials that happen within 50ms.

For excitatory synapses, where the presynaptic neuron is an excitatory neuron, plasticity was facilitated using the rules observed by Bi and Poo (1998). When the presynaptic neuron fires before the postsynaptic neuron, where *dt = t_pre - t_post < 0*, we have causal firing, and we increase the number of AMPA receptors using the following formula:

$$\Delta R\_AMPA = (1*10^7) \times f(R\_AMPA) \times exp(-\Delta t / tau)$$

When the postsynaptic neuron fires before the presynaptic neuron, where *dt = t_pre - t_post > 0*, we have acausal firing, and we decrease the number of AMPA receptors using the following formula:

$$\Delta R\_AMPA = (-1*10^7) \times f(R\_AMPA) \times exp(\Delta t / tau)$$

For inhibitory synapses, we also implemented similar logic following experimental observations in the entorhinal cortex by Haas et al. (2006). When the presynaptic neuron fires before the postsynaptic neuron, we have causal firing, and we increase the number of GABA receptors using the following formula:

$$\Delta R\_GABA = (1*10^7) \times f(R\_GABA) \times exp(\Delta t / tau)$$

When the postsynaptic neuron fires before the presynaptic neuron, we have acausal firing, and we decrease the number of GABA receptors using the following formula:

$$\Delta R\_GABA = (-1*10^7) \times f(R\_GABA) \times exp(-\Delta t / tau)$$

To prevent over-strengthening synapses, we implemented soft normalization through f(R), which takes the value *1/(R-1) for R>5*, and 0.25 otherwise. This creates a stabilizing negative feedback where it gets progressively harder for a strong synapse to get even stronger. We also implemented hard bounds, with *0 < R < 2×R_initial*. *tau* was chosen as 4e-3.

### 4) Network Architecture and Simulation

The network consisted of 200 neurons, 40 inhibitory and 160 excitatory. Each neuron had an 80% chance to create a connection with any other neuron, creating around 32000 total synapses. Receptor counts were initialized using Gaussian distributions, with AMPA having 120 mean and 12 variance, and GABA having 200 mean and 6 variance. These values were chosen to achieve sparse activation of the network, though other combinations are possible.

To initiate network activity, an external current of 80 microamps per neuron was injected from simulated patch clamps to 30 randomly selected excitatory neurons for 200ms, with a mean start time of 400ms after simulation start. After this initial stimulation, no external stimulus was given to the system.

Simulations used a timestep of 10 microseconds, meaning 100,000 steps were required for each second of simulation. Membrane voltages were recorded every 100 steps (1ms) for each neuron, with 2 runs conducted for 500 seconds, and 1 run being 1800 seconds long to analyze self-sustainability on longer timescales.

The simulation was implemented using C++/CUDA on an RTX 4060 Mobile GPU, with the 500-second simulations taking approximately 4 GPU hours, and the 1800-second simulation taking approximately 14 GPU hours. Complete parameter values are provided in Table 1 in the Appendix.

## IV. Experiments and Results

We conducted three simulations: two 500 s runs (Runs 1 and 2) and one 1800 s run (Run 3). In each case, the network comprised 200 Hodgkin-Huxley-based neurons, of which 160 were excitatory and 40 inhibitory. Synaptic connectivity was generated probabilistically, with each neuron having a 0.8 probability of forming a connection with another neuron, yielding approximately 32,000 synapses in total. In every simulation, 30 randomly selected excitatory neurons were given a brief 200 ms stimulus within the first second, after which no further external input was applied. Run 3 was used to characterize long-term network dynamics, while Runs 1 and 2 served as shorter independent realizations to support the findings from the 1800 s simulation and to evaluate the consistency of the observed behavior.

As a first step in the analysis, we quantified the mean firing rate of each neuron over the entire simulation period (Figure 1). After a brief 200 ms initialization stimulus to 30 neurons, no further external input was applied. Each bar in Figure 1 represents one neuron (x-axis: neuron ID 0-199). Despite the high level of network connectivity, activity remained sparse. Only 3% of neurons displayed average firing rates greater than 5 Hz, whereas 30% fired between 1 and 5 Hz, and 67% remained below 1 Hz. The overall mean firing rate across the network was 1.13±1.34Hz. Detailed statistics for all runs can be found in Table 1.

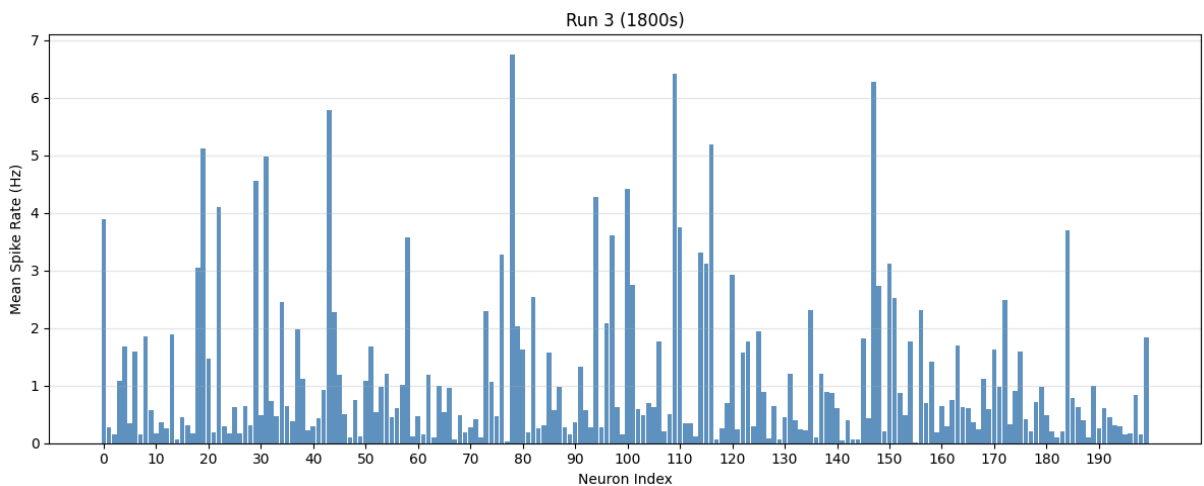

*Figure 1. Distribution of mean firing rates across all 200 neurons during the full 1800 s simulation Run 3.*

|        | Run 1       | Run 2       | Run 3       |
|--------|-------------|-------------|-------------|
| <1Hz   | 44.5%       | 65.5%       | 67%         |
| 1-5Hz  | 45.5%       | 27%         | 30%         |
| >5Hz   | 10%         | 7.5%        | 3%          |
| Mean±std | 2.06Hz±2.33 | 1.42Hz±2.14 | 1.13Hz±1.34 |

*Table 1. Neuron spike rate distribution in the population at different runs.*

Figure 2 presents raster plots of the simulated activity, with each spike shown as a point indexed by time and neuron identity. Time is shown on the x-axis and neuron identity on the y-axis (0-39 inhibitory; 40-199 excitatory). Following a brief 200 ms stimulus delivered to 30 randomly selected neurons in the first second, the network continued to generate activity autonomously. These plots indicate that the network remains active over the full simulation interval. Notably, after the initial stimulus, the network does not relax into a single fixed firing rhythm. Rather, it evolves into a semistable dynamical regime in which some neurons fire relatively regularly while others remain temporally variable. This regime is not static: at approximately 190 s, the network exhibits an abrupt transition in its collective rhythm, accompanied by rapid changes in the firing rates of many neurons despite the absence of any external perturbation. Comparable spontaneous metastable-like transitions are visible as abrupt reorganizations of the firing pattern, evident, for example, near t ≈ 650s and 920s. The network maintained sparse, irregular firing throughout the full 30 min interval in the absence of an ongoing external drive, indicating self-sustained activity.

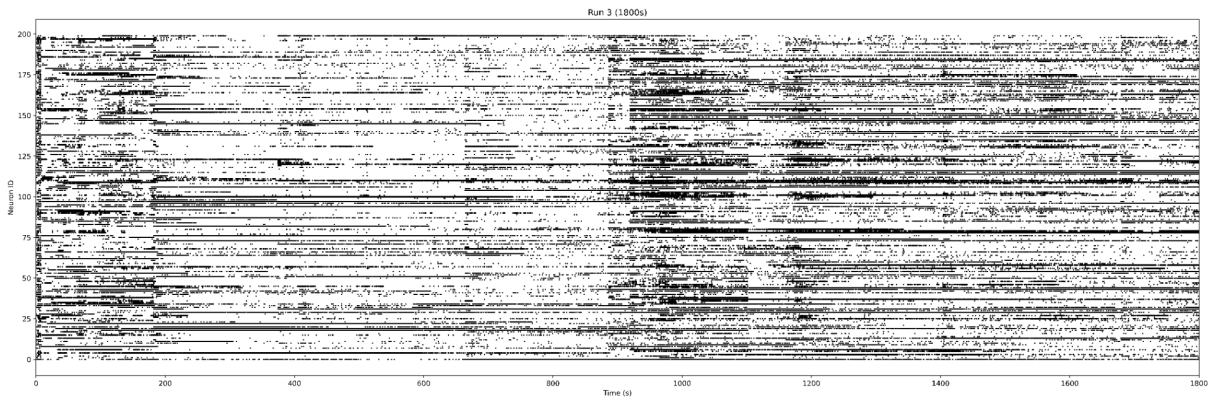

*Figure 2. Raster plot of all spikes generated by the 200-neuron network over the 1800 s simulation of Run 3.*

To quantify how broadly activity was distributed across the population, we measured participation rates, defined as the fraction of neurons generating at least one spike within a specified time window (Figure 3; Table 2). Participation increased strongly with the duration of the observation window. In Run 3, mean participation rose from 24.52±9.47% at 1s to 50.42±14.26% at 10s to 71.06±14.69% at 50s (see Table 2 for all runs). Thus, although firing was sparse at the single-neuron level, the majority of neurons participated in the network dynamics when activity was integrated over longer temporal scales. This pattern reflects temporally sparse activity: only a minority of neurons (~25%) are active at any given second, yet most neurons (>70%) contribute over longer temporal horizons.

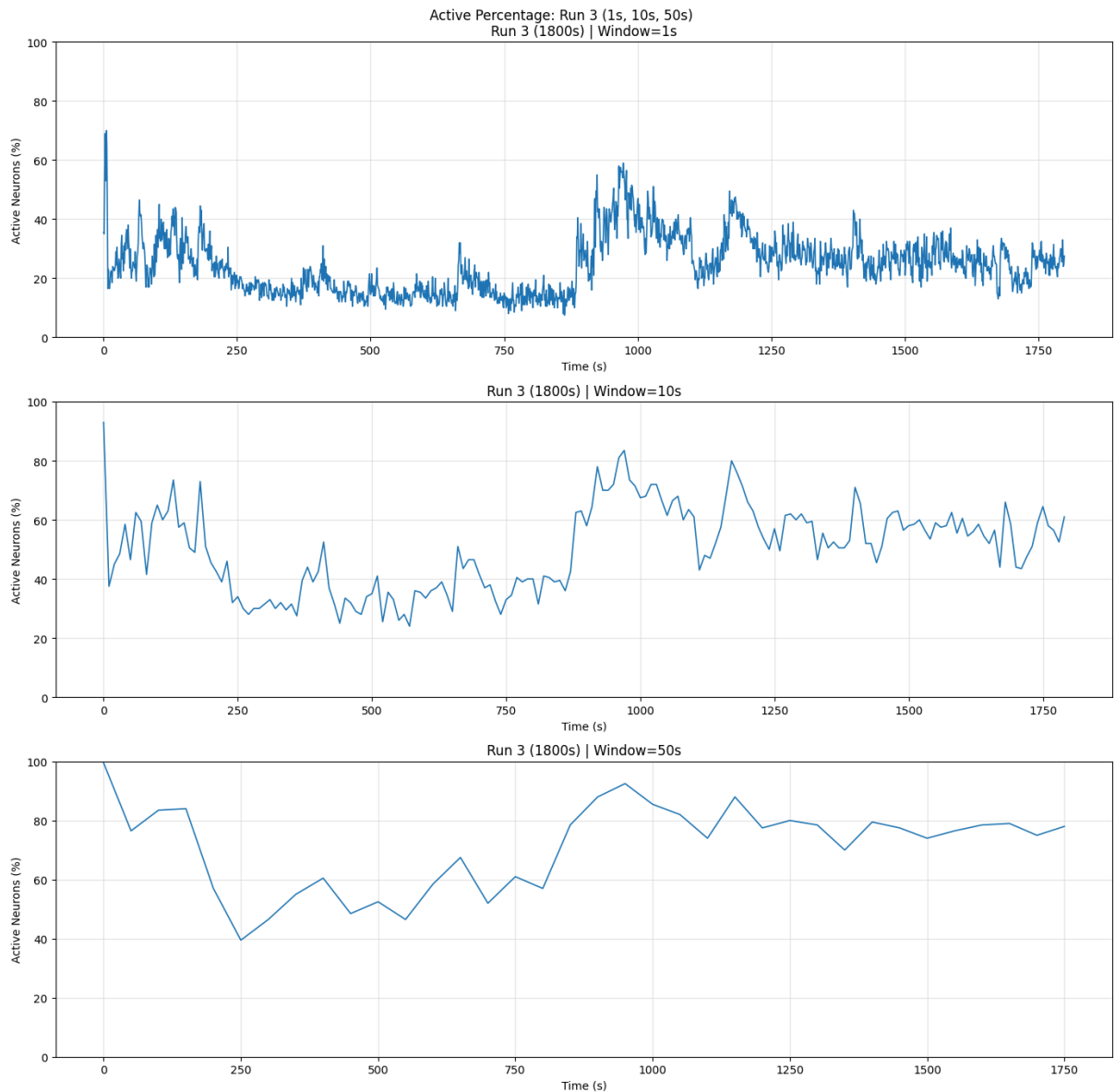

*Figure 3. Fraction of neurons generating at least one spike within windows of different durations (1 s, 10 s, and 50 s) over the 1800 s simulation at Run 3.*

|            | Run 1         | Run 2          | Run 3          |
|------------|---------------|----------------|----------------|
| 1s Window  | 42.48±9.49 %  | 29.41±10.05 %  | 24.52±9.47 %   |
| 10s Window | 71.85±8.05 %  | 55.98±14.97 %  | 50.42±14.26 %  |
| 50s Window | 87.80±6.88 %  | 74.25±16.32 %  | 71.06±14.69 %  |

*Table 2 Participation rates, defined as the percentage of neurons that had at least one spike within a given time window.*

Irregularity in spike timing was further examined using the Fano factor, calculated as the variance-to-mean ratio of spike counts. In Figure 4, the orange line denotes the population means, and the shaded band indicates the full range across neurons. Population-mean values remained between 1 and 2 over the course of the simulation, consistent with sustained irregular firing and approximately

Poisson-like variability. At the single-neuron level, firing statistics were heterogeneous, ranging from relatively regular (F<0.5) to approximately Poisson (F≈1), to strongly bursty (F>10). This sustained irregularity is consistent with asynchronous irregular dynamics observed in cortical networks.

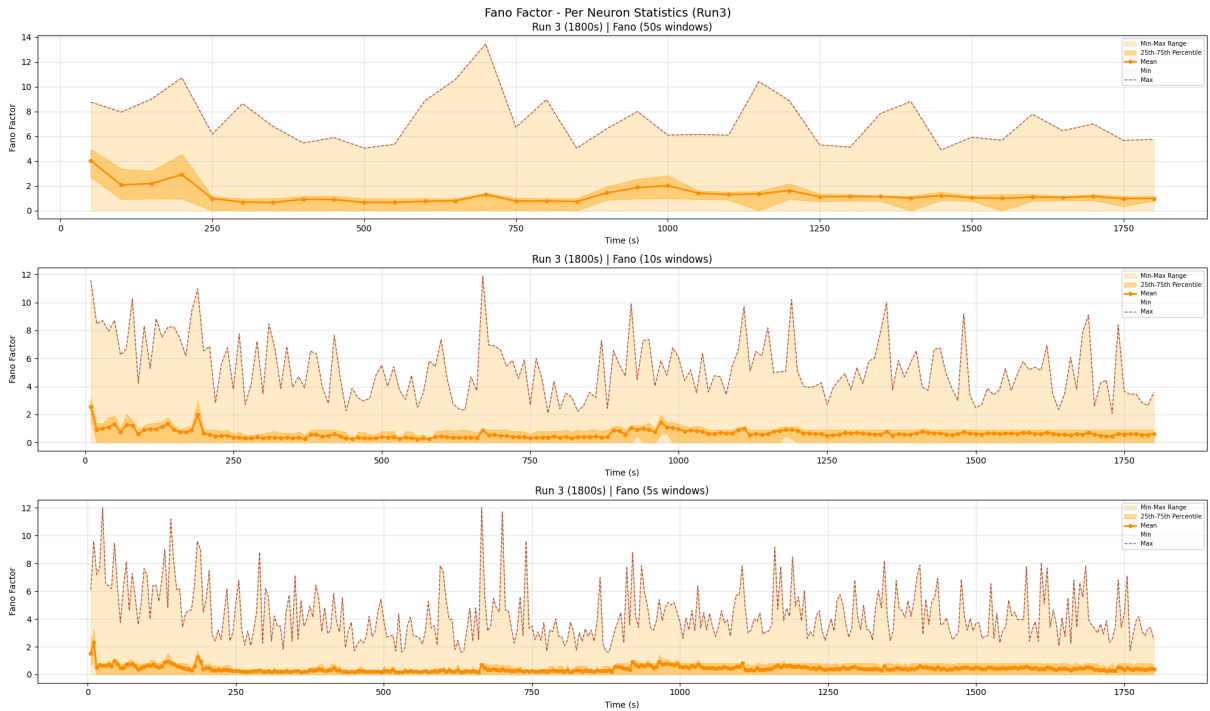

*Figure 4. Temporal evolution of the Fano factor for Run 3, computed from spike counts in sliding windows of 50 s, 10 s, and 5 s.*

## V. Discussion

A biologically realistic model of the nervous system should capture not only the structural and mechanistic features of neural tissue, but also its capacity for autonomous persistence. In neural systems, this requirement necessarily extends to stochasticity and synaptic plasticity, both of which are fundamental to real biological function. The present study was therefore designed to examine whether a biologically grounded neuronal population can sustain its own activity in a digital environment and, if so, how that activity evolves under the combined influence of recurrent connectivity, refractory dynamics, stochasticity, and plasticity.

The results show that a highly connected neuronal population can indeed maintain persistent activity without ongoing external stimulation. In our simulations, a brief 200 ms perturbation delivered to only 15% of the neurons during the initial phase of the run was sufficient to induce continuous activity that persisted throughout the full simulation window. Importantly, this activity neither saturated nor collapsed into silence. These findings suggest that self-sustained activity can arise as an intrinsic property of the network when the excitatory-inhibitory balance and recurrent organization are appropriately configured. In this respect, the present work adds to the relatively limited literature explicitly addressing autonomous self-sustained neural dynamics.

A major contribution of the model lies in its high degree of biological grounding. The excitatory-inhibitory ratio was chosen to reflect cortical organization, excitatory transmission was mediated through AMPA receptors, inhibitory signaling through GABAergic mechanisms, and synaptic adaptation was implemented through STDP. Receptor numbers and spontaneous vesicular release dynamics were constrained by empirical findings, and action potentials were modeled using the Hodgkin-Huxley formalism. Moreover, all membrane dynamics were simulated continuously at high temporal resolution. This is important because the irregularity observed in biological firing is difficult to reproduce with simplified, low-resolution, threshold-reset models alone. The use of continuous Hodgkin-Huxley dynamics therefore strengthens the physiological plausibility of the observed self-sustained regime.

The present findings also support the idea that excitatory-inhibitory balance alone may not be sufficient to maintain stable autonomous activity. Previous studies have suggested that persistent activity without external input requires some form of effective synaptic suppression. Physiologically, such suppression resembles the influence of a longer refractory recovery, which is naturally represented by the Hodgkin-Huxley framework. Our results are consistent with this interpretation: networks in which action potentials were modeled with Hodgkin-Huxley dynamics were capable of sustaining activity over extended durations while remaining irregular and non-saturating. This suggests that realistic refractory constraints may be one of the critical ingredients required for self-sustainability.

Connectivity strength also played a central role in shaping the observed dynamics. At the approximate scale of a cortical microcolumn, dense recurrent connectivity is biologically plausible, and in our model, such connectivity supported both continued activity and ongoing synaptic modification through STDP. At the same time, dense connectivity did not lead to globally elevated firing rates. Despite an 80% connection probability, the network remained sparse, with most neurons firing below 5 Hz and only a small fraction reaching higher rates. This shows that sparse coding can emerge naturally from the internal dynamics of the network rather than from imposed sparsity penalties or external regulation. The most likely stabilizing mechanism is the recurrent interaction between excitation and inhibition: as network activity increases, inhibitory recruitment also increases, thereby limiting runaway excitation.

The system also displayed a strongly stochastic mode of operation. Spike counts across different time windows were broadly consistent with Poisson-like irregularity, and both single-neuron and population-level measures showed persistent variability over time. Different random initializations led to substantially different detailed trajectories, but the main macroscopic properties of the network—self-sustainability, sparsity, stochasticity, and metastability—remained intact across runs. This suggests that stochasticity is not an artifact of a particular seed or configuration, but rather a robust emergent property of the underlying biological mechanisms.

An especially important observation is that this stochasticity persisted in the absence of external input. Prior work has often examined irregular activity in networks that remain at least partially input-driven, making it difficult to distinguish intrinsic variability from externally sustained fluctuation. In the present model, by contrast, autonomous activity was maintained entirely through internal dynamics after a very brief initial perturbation. This makes the system particularly relevant for understanding how spontaneous neural activity may persist in living tissue and how internally generated dynamics may preserve, transform, or obscure traces of prior inputs.

The activity was not merely random, however. The network repeatedly exhibited metastable transitions, with spontaneous shifts between partially distinct operating regimes. These transitions occurred without external intervention, indicating that the system continuously reorganized itself while preserving its overall functional state.

Taken together, these findings point to a deeper organizational principle: degeneracy. Multiple distinct microscopic configurations were able to produce the same major macroscopic properties. In other words, the network's core functional behavior did not depend on exact initial synaptic weights or on a single privileged trajectory. Instead, the persistence of self-sustainability, sparse coding, stochasticity, and metastability appears to emerge from broader system-level principles, including recurrent structure, membrane biophysics, plasticity, and excitatory-inhibitory balance. Such degeneracy is a desirable feature for biological systems because it confers robustness: the system can preserve function while tolerating internal variation and external perturbation. At the same time, it allows exploration of multiple internal states without loss of overall operational coherence.

## VI. Conclusion, Limitations and Future Directions

This study demonstrates that a biologically grounded recurrent neuronal population can sustain autonomous activity for extended periods after only a brief initial perturbation and without any subsequent external input. The observed activity remained sparse, irregular, metastable, and non-saturating despite dense connectivity, indicating that self-sustained firing can arise as an intrinsic emergent property of the system rather than as a transient response to continued stimulation.

By combining Hodgkin-Huxley membrane dynamics, STDP-based synaptic plasticity, biologically informed synaptic parameters, and explicitly modeled intrinsic stochasticity, the present work provides a more realistic in silico framework for studying persistent neural activity than is possible with simplified threshold-reset models alone. The results further suggest that realistic refractory dynamics, strong recurrent connectivity, and a stable excitatory-inhibitory balance jointly enable long-duration self-sustainability, while stochastic biological mechanisms preserve irregularity and dynamical flexibility.

The findings also show that self-sustained activity in such a system is not rigidly tied to a single microscopic configuration. Instead, core network properties were preserved across distinct random initializations and repeated metastable transitions, indicating a degenerate and therefore robust operating regime. This robustness is likely to be essential for biological neural systems, which must maintain function despite variability, noise, and changing internal conditions.

More broadly, the study supports the view that spontaneous neural activity, sparse coding, stochasticity, and plastic adaptation are not separate phenomena, but interdependent features of living neural organization. Modeling them together is therefore essential for understanding how neural systems preserve activity, sustain computation, and potentially retain traces of prior input over time. In this sense, the present framework offers a useful platform for future investigations of self-sustained dynamics, memory trace formation, and inverse inference in noisy biological neural networks.

Several limitations of the present study should be noted. First, the network comprised only 200 neurons and therefore remains far smaller than real neural circuits. The degree to which the central findings generalize to larger populations has not yet been established. Second, while the model displayed biologically plausible dynamics and features that may support efficient information storage and processing, its computational capabilities were not directly evaluated in task-oriented settings. Third, the current implementation omits several biologically important complexities, including multiple neurotransmitter subclasses and multicompartment neuronal structure.

Future studies should address these limitations by scaling the model to larger populations, incorporating more detailed biological mechanisms, and evaluating performance in explicit computational tasks involving structured external inputs. Such developments may ultimately support new applications in artificial intelligence and neuroscience, particularly on energy-efficient neuromorphic chips and FPGA-based systems.

Notwithstanding these limitations, the present work shows that a simple, highly connected Hodgkin-Huxley-based network can exhibit robust self-sustainability and adaptive dynamics over a prolonged 30-minute time window, while retaining properties of potential computational value.

# VII. Appendix

$$\Delta u = 10^3 (u - u_{rest})$$

$$\phi = 3^{\frac{T-6.3}{10}}$$

$$\alpha_m = \phi \frac{(\epsilon + 2.5 - 0.1\Delta u)}{(\epsilon + \exp(2.5 - 0.1\Delta u) - 1)}$$

$$\alpha_n = \phi \frac{(\epsilon + 0.1 - 0.01\Delta u)}{(\epsilon + \exp(1 - 0.1\Delta u) - 1)}$$

$$\alpha_h = 0.07\phi \exp\left(-\frac{\Delta u}{20}\right)$$

$$\beta_m = 4\phi \exp\left(-\frac{\Delta u}{18}\right)$$

$$\beta_n = 0.125\phi \exp\left(-\frac{\Delta u}{80}\right)$$

$$\beta_h = \frac{\phi}{\exp(3.0 - 0.1\Delta u) + 1}$$

| Parameter | Symbol | Value | Description |
| --- | --- | --- | --- |
| Resting membrane voltage | $u_{rest}$ | -65 mV | Resting membrane voltage |
| Threshold membrane voltage | $u_{thres}$ | -35 mV | Voltage threshold for spikes |
| Max membrane voltage | $u_{max}$ | 0.7 V | Hard limit on maximum membrane voltage |
| Min membrane voltage | $u_{min}$ | -0.1 V | Hard limit on minimum membrane voltage |
| Inhibitory threshold | $thres_{inh}$ | -70mV | The threshold below which inhibitory currents are attenuated |
| Inhibitory decay rate | $\alpha_{decay}$ | 500/mV | Attenuation factors for inhibitory currents below the inhibitory threshold |
| Sodium equilibrium voltage | $E_{Na}$ | 50 mV | The potential where the net Na current is zero. |
| Potassium equilibrium voltage | $E_K$ | -77 mV | The potential where the net K current is zero. |
| Leakage equilibrium voltage | $E_L$ | -60 mV | The reversal potential for the leak current. |
| Max sodium conductance | $g_{Na_{max}}$ | 120 mS | Maximum possible conductance when all Na channels are open. |
| Max potassium conductance | $g_{K_{max}}$ | 50 mS | Maximum possible conductance when all K channels are open. |
| Max leakage conductance | $g_{L_{max}}$ | 0.3 mS | Constant conductance of the non-specific leak channels. |
| Membrane capacitance | $C_m$ | 0.1 μF | The capacity of the lipid bilayer to store charge. |
| Temperature | $T$ | 20 C | The temperature measured in the neuron pool |
| m rate coefficient | $\tau_m$ | 0.3 | Time adjustment factor for m |
| n rate coefficient | $\tau_n$ | 0.32 | Time adjustment factor for n |
| h rate coefficient | $\tau_h$ | 0.6 | Time adjustment factor for h |

*Table 3. Default neuron parameters*

| Parameter | Symbol | Value |
|---|---|---|
| STDP time window | $dt_{STDP}$ | 50ms |
| STDP time constant | $\tau$ | 4ms |
| Vesicle Voltage Change Factor | $du_{per-ves}$ | 1/150 |
| Synaptic update period | $T_{update}$ | 1 ms |
| Maximum synapse current | $I_{syn_{max}}$ | 40 μ |
| AP release probability ) | $p_{AP\,release}$ | 0.5 |
| Spontaneous AP release probability | $p_{spn\,AP\,release}$ | 0.001 |
| Lookback AP period | $T_{lookback_{AP}}$ | 100 ms |
| Vesicle release base period | $T_{ves\,release\,base}$ | 50 ms |
| Vesicle release decay rate | $\alpha_{decay}$ | 100/s |

*Table 4. Default synapse parameters*